\documentclass[conference]{IEEEtran}
\usepackage{amsmath,amsfonts}
\usepackage{algorithm,algpseudocode}
\usepackage{array}
\usepackage{multirow}
\usepackage{caption}
\usepackage{subcaption}
\usepackage{graphicx}
\usepackage{textcomp}
\usepackage{stfloats}
\usepackage{url}
\usepackage{verbatim}
\usepackage{graphicx}
\usepackage{cite}
\usepackage{booktabs}
\usepackage{bm}
\usepackage{color}
\usepackage{hyperref}
\usepackage{enumitem}
\usepackage{siunitx}
\usepackage{makecell}
\usepackage{siunitx} % For handling units

\newcommand{\titlename}{BridgeGen}

\begin{document}

\title{\titlename: Bridging Data-Driven and Knowledge-Driven Approaches for Safety-Critical Scenario Generation in Automated Vehicle Validation}

\author{
    \IEEEauthorblockN{
        Kunkun Hao\textsuperscript{1},
        Lu Liu\textsuperscript{1,4},
        Wen Cui\textsuperscript{1,2,3},
        Jianxing Zhang\textsuperscript{1},\\
        Songyang Yan\textsuperscript{1,4},
        Yuxi Pan\textsuperscript{1},
        and Zijiang Yang\textsuperscript{1,4}
    }
    
    \vspace{0.3cm}
    
    \IEEEauthorblockA{
        \textsuperscript{1}Research Center of Synkrotron, Inc., Xi'an, China\\
        \{haokunkun, liulu, zhangjianxing, tayyim, yuxip, yang\}@synkrotron.ai
    }
    
    \vspace{0.3cm}
    
    \IEEEauthorblockA{
        \textsuperscript{2}Institute for Interdisciplinary Information Core Technology, Xi'an, China\\
        \textsuperscript{3}Academy of Advanced Interdisciplinary Research, Xidian University, Xi'an, China\\
        \{cuiw\}@iiisct.com
    }
    
    \vspace{0.3cm}
    
%    \IEEEauthorblockA{
%        \textsuperscript{4}AI Laboratory, Chongqing Changan Automobile Co. Ltd, Chongqing, China\\
%        \{luoyg3\}@changan.com.cn
%    }

    \vspace{0.3cm}
    
    \IEEEauthorblockA{
        \textsuperscript{4}Department of Computer Science and Technology, Xi'an Jiaotong University, Xi'an, China\\
        \{liulu615, tayyin\}@stu.xjtu.edu.cn
    }
}

\maketitle

\begin{abstract}
Automated driving vehicles~(ADV) promise to enhance driving efficiency and safety, yet they face intricate challenges in safety-critical scenarios. As a result, validating ADV within generated safety-critical scenarios is essential for both development and performance evaluations. This paper investigates the complexities of employing two major scenario-generation solutions: data-driven and knowledge-driven methods. Data-driven methods derive scenarios from recorded datasets, efficiently generating scenarios by altering the existing behavior or trajectories of traffic participants but often falling short in considering ADV perception; knowledge-driven methods provide effective coverage through expert-designed rules, but they may lead to inefficiency in generating safety-critical scenarios within that coverage. To overcome these challenges, we introduce \titlename, a safety-critical scenario generation framework, designed to bridge the benefits of both methodologies. Specifically, by utilizing ontology-based techniques, \titlename\ models the five scenario layers in the operational design domain~(ODD) from knowledge-driven methods, ensuring broad coverage, and incorporating data-driven strategies to efficiently generate safety-critical scenarios. An optimized scenario generation toolkit is developed within \titlename. This expedites the crafting of safety-critical scenarios through a combination of traditional optimization and reinforcement learning schemes. Extensive experiments conducted using Carla simulator demonstrate the effectiveness of \titlename\ in generating diverse safety-critical scenarios.
\end{abstract}

\section{Introduction}

Automated driving vehicles~(ADV) are dedicated to improving travel efficiency and reducing traffic accidents, but the actual implementation of autonomous driving still needs to address some key issues. First, autonomous driving systems are composed of various AI algorithm models, characterized by complexity and uninterpretability; and the driving environment faced by autonomous vehicles is incredibly complex, with endless unknown long-tail scenarios in the real world~\cite{zhao2021comparative}. These scenarios may cause safety problems for autonomous vehicles, so they must be thoroughly validated before deployment~\cite{li2020theoretical, ma2022verification}. Public road testing, proving ground testing, and simulation testing are the three pillars of ADV validation. Among these, simulation testing stands out as the advantages of high efficiency, low cost, and reproducibility~\cite{zapridou2020runtime}, while scenario-based simulation testing is the most effective method~\cite{sun2021scenario}. It aims to automate and efficiently generate a variety of safety-critical scenarios within the operational design domain~(ODD)~\cite{thorn2018framework}, fulfilling the SOTIF~(safety of the intended functionality) testing requirements~\cite{iso21448}.

% figure 1
\begin{figure}
    \centering
    \includegraphics[width=\linewidth]{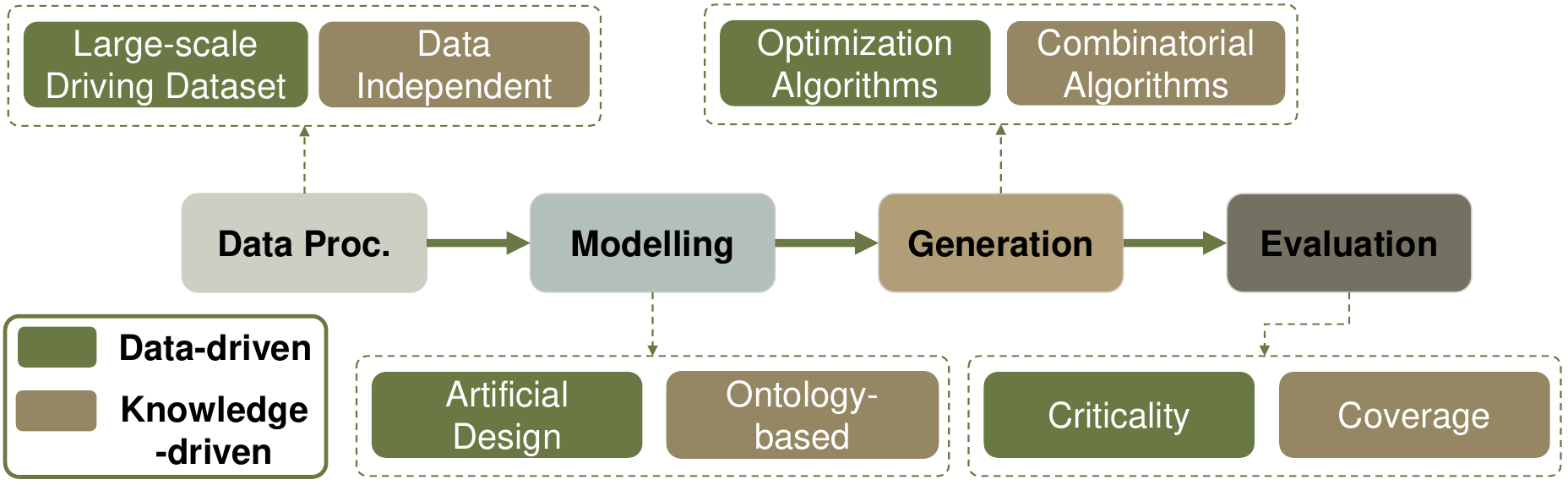}
    \caption{{\bf Comparison between data-driven and knowledge-driven scenario generations.} A general scenario generation pipeline consists of four main components: data processing, scenario modeling, scenario generation, and scenario evaluation. Data-driven and knowledge-driven solutions have their own unique characters in each of the four components.}
    \label{fig:workflow}
\end{figure}

Scenario-based testing can generally be categorized into two methods: data-driven and knowledge-driven~\cite{cai2022survey, zipfl2023comprehensive}, as illustrated in Fig.~\ref{fig:workflow}. Data-driven methods, which utilize datasets either from real-world recordings or simulation platforms and often employ advanced reinforcement learning algorithms to generate safety-critical scenarios~\cite{feng2021intelligent}. In particular, these algorithms can effectively extract details from the behavior or trajectory distributions within the dataset, and then derive scenarios by modifying existing behaviors or trajectories. The algorithm can be further fine-tuned using feedback from various criticality metrics~\cite{westhofen2023criticality,xu2021accelerated}. However, a common limitation of most current methods is their primary focus on the validation of planning-level and control-level modules. They often neglect the validation of the perception-level module, even though variations in perception-level results could directly impact subsequent modules. \textcolor{blue}{The growing trend of addressing perception-level limitations has yet to be implemented across other modules, limiting overall usability~\cite{fremont2019scenic,fremont2022scenic}.} On the other hand, knowledge-driven methods, such as ontology-based methods~\cite{de2022towards, armand2014ontology, kohlhaas2014semantic}, employ expert experience or traffic rules to model the ODD, which consists of a 5-layer structure~\cite{bagschik2018ontology}. This approach covers all elements in perception-level, planning-level, and control-level scenarios. However, the search for the exact value of an element, e.g., rain levels, width or length of lanes, and the position of traffic participants, in generating scenarios relies on burdensome brute-force search or combinatorial testing methods~\cite{westhofen2022using, wotawa2020ontology}, resulting in inefficiency. Given these observations, we argue that it is crucial to combine the advantages of both data-driven and knowledge-driven methods. By doing this, we can efficiently generate safety-critical while ensuring comprehensive coverage across all 5 layers of the ODD.

\begin{figure*}
\centering
\includegraphics[width=.7\linewidth]{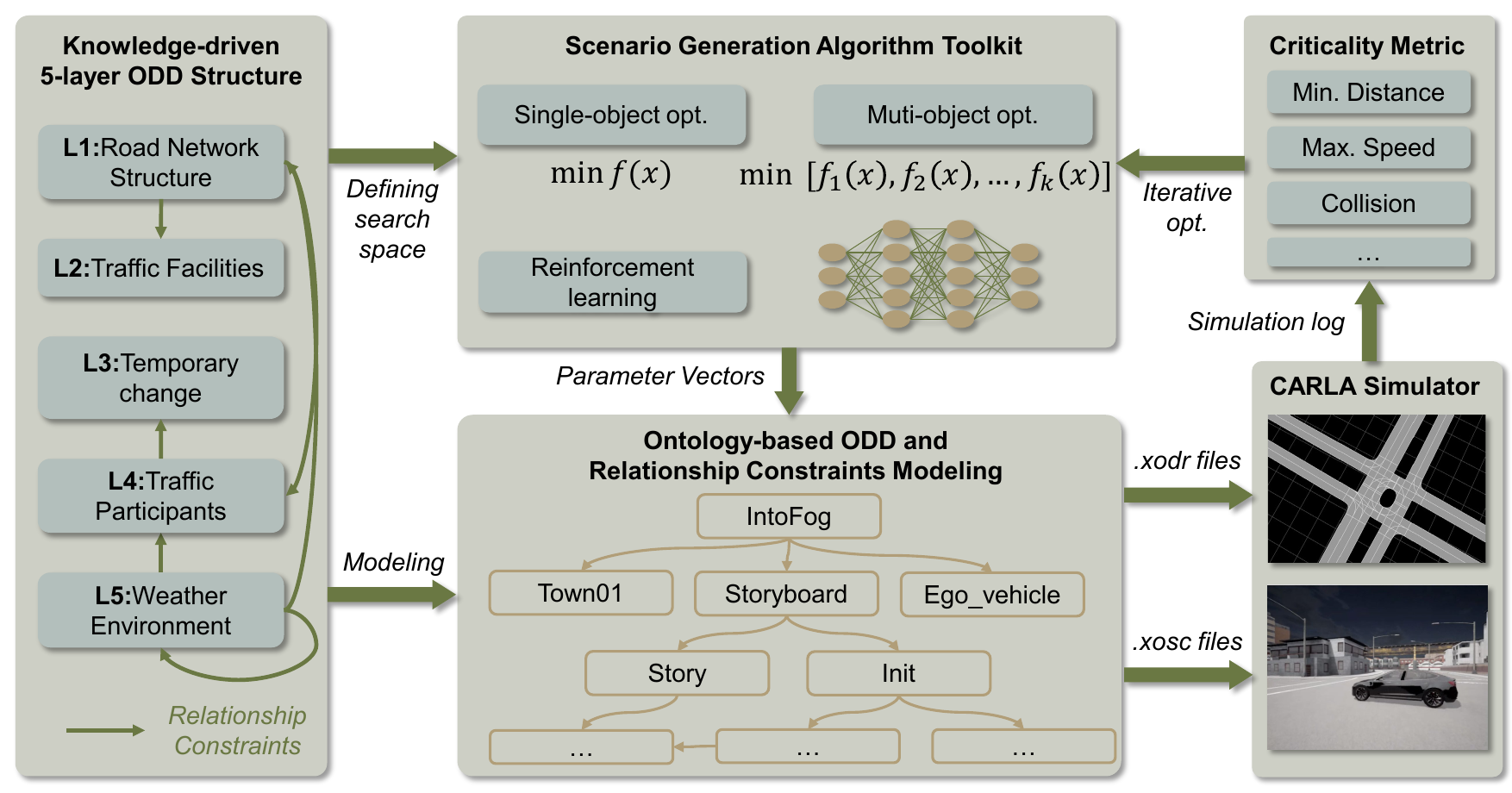}%
\caption{The overall framework of \titlename.}
\label{fig:frame}
\end{figure*}

In this paper, we propose \titlename, a solution for automated vehicle validation. To the best of our knowledge, this is the first safety-critical scenario generation solution that effectively bridges both data-driven and knowledge-driven approaches. We draw the overall framework in Fig.~\ref{fig:frame}, and outline our contributions as follows.
\begin{enumerate}
    \item To ensure comprehensive coverage, we employ ontology-based methods to perform semantic analysis and modeling of the 5-layer scenarios in the ODD, including both dynamic and static elements. To guarantee the realism of the generated scenarios, we also model relationship constraints across different layers of scenario elements, such as not having dry roads in rainy weather or strong winds in dense fog. Further, \titlename\ is designed to be ready for common simulators by generating OpenScenario description files for dynamic traffic facilities and OpenDrive files for static road network structures.
    \item To boost the efficiency in generating safety-critical scenarios, and to avoid burdensome brute-force knowledge reasoning or combinatorial testing, we have developed an optimization-based scenario generation toolkit. This toolkit includes traditional optimization search algorithms and advanced reinforcement learning algorithms. Moreover, the quality of chosen evaluation metrics directly determines the effectiveness of critical scenario generation. Therefore, to better guide the optimization direction of the scenario generation algorithm, we provide a criticality metric configuration module. This module allows convenient configurations of the scenario generation algorithm with interest metrics per testing task, such as the minimum distance of traffic participants, the corresponding speed of the ego vehicle at the minimum distance, whether there is a collision, etc. 
    \item Finally, we evaluated the performance of \titlename\ against traditional safety-critical scenario generation solutions using the Carla simulator. The experimental results reveal that \titlename\ can efficiently generate a large variety of safety-critical scenarios. \titlename\ also facilitates rapid comparative verification of different generation algorithms, thereby accelerating follow-up research for scenario generation algorithms.
\end{enumerate}

\section{Related Work}

We summarize recent progress in generating safety-critical scenarios for ADV validation in this section.

\subsection{Knowledge-driven solutions.}

Knowledge-driven solutions typically utilize expert experience to analyze and model elements across different ODD layers. Initially, the absence of a universal scenario structure led to generated scenarios that were often confusing and poorly constructed. To address this issue, researchers introduced the philosophical concept of ontology into the design procedures~\cite{gruber1993translation}. The goal was to formalize knowledge in a general and well-structured manner, thereby generating explainable scenarios for both humans and machines~\cite{hummel2008scene}. Focusing first on road structures, more recent studies are generating scenarios from different ODD layers using ontology~\cite{de2022towards, armand2014ontology, kohlhaas2014semantic}. Moreover, there is a growing trend to employ ontology-based methods to apply these modeled scenarios to simulations, expediting the validation of ADV~\cite{chen2018ontology,bogdoll2022one,tahir2021intersection,li2020ontology,westhofen2022using,huang2019ontology}. Although promising, to maintain good coverage, existing studies employ brute-force search or combinatorial testing methods. These methods may lead to inefficient and time-consuming reasoning, and a combinatorial explosion of possibilities. Moreover, due to the high cost of these methods, recent work tends to focus on specific layers of ODD, weakening its ability to cover all 5-layer ODD scenarios or represent the natural interplay between scenario elements. In contrast, to ensure efficiency, \titlename\ leverages both traditional optimization and advanced learning techniques in generating safety-critical scenarios. This enables \titlename\ to comprehensively model all five layers of ODD and cover the constraint relationships between elements.

\subsection{Data-driven solutions.}
ADV datasets, from simulation platforms or real-world recordings, are rich resources of diverse driving scenarios, among which safety-critical are particularly valuable for ADV validation. By formulating optimization problems with critical metrics as objectives, the safety-critical scenario can be generated using existing genetic or other heuristic search algorithms~\cite{zhou2022genetic,moghadam2022machine,lu2023deepscenario, lu2022learning,hanselmann2022king}. Further generalization can be achieved by altering the existing behavior or trajectories of traffic participants.  Additionally, adversarial models employing reinforcement learning techniques can be trained on these datasets, providing generators of natural safety-critical scenarios~\cite{rempe2022generating,chen2017end,santana2016learning,pan2017virtual}. However, most datasets contain scenario details processed after the perception stage of ADV, thus focusing predominantly on the planning level and control level. Consequently, current studies lack an investigation of the impact of perception-level modules on safety-critical scenarios. This oversight can diminish the coverage of generated safety-critical scenarios, as changes in perception-level results may directly influence subsequent modules. Instead, by incorporating the effects of perception-level modules, \titlename\ can generate safety-critical scenarios that span all levels of modules, from perception to planning and control, ensuring broader coverage and more robust validation.
% figure 2

% figure 3
\begin{figure}
    \centering
    \includegraphics[width=.9\linewidth]{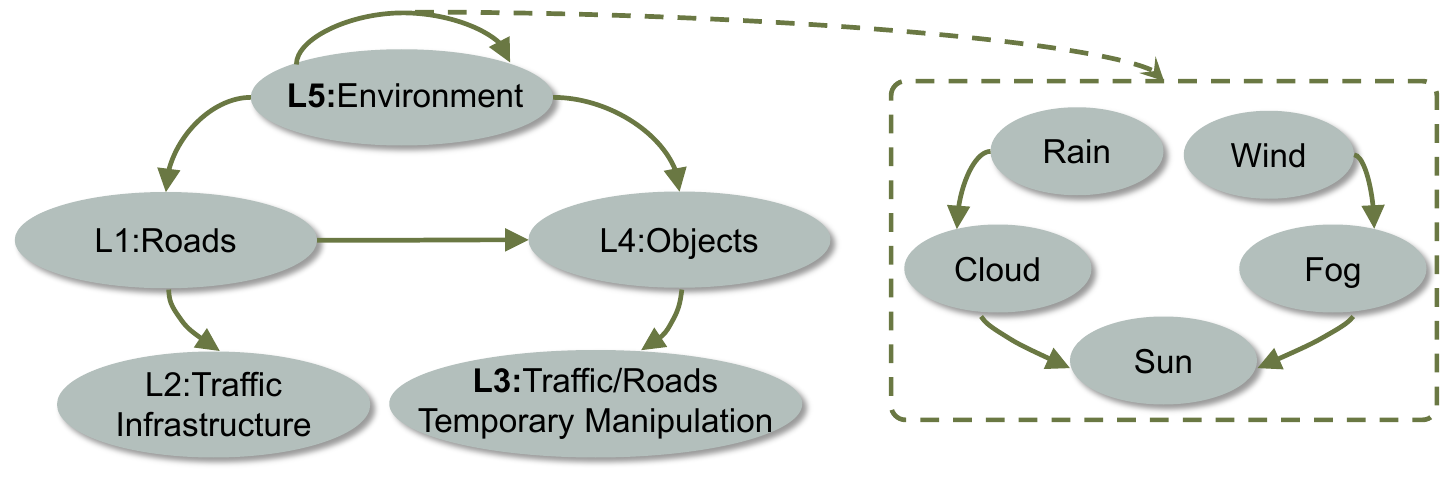}
    \caption{ODD scenario elements constraints.}
    \label{fig:Constraint relationship}
\end{figure}

\section{Search Space Construction and Constraints on Typical ODD Scenario Elements}

The operational design domain~(ODD) was proposed in prior studies to formalize the elements present in ADV testing scenarios~\cite{czarnecki2018operational,hillen2020model}. In this section, we first define the search space corresponding to these elements. Subsequently, we delve deeper to investigate the inherent constraints among these elements, ensuring the realism of the generated scenario.

\subsection{Search Space Construction}\label{sec:Search Space Construction}
ODD typically categorizes scenario elements into five layers: Layer1 for the road network structure, Layer2 for the traffic facilities, Layer3 for the temporary change, Layer4 for the traffic participants, and Layer5 for the weather environment. In detail, the road network structure includes attributes like lane width, number of lanes, lane length, and intersection size. Traffic facilities include elements like traffic lights, signs, and lane markings. Temporary changes in the elements can reflect situations such as traffic accidents or congestion. The traffic participants layer captures the behaviors and events in the scenario. The weather environment layer includes conditions like rain, snow, wind, lighting, and road states. %The complexity of the structure intensifies from the lower to the higher layers.

To effectively model and instantiate scenarios, it is crucial to define a search space that aligns with the ODD's five-layer model, confining the search space in the process. Within this confined space, we can derive critical search vectors, which define the elements of the generated scenarios. Specifically, we group the layers into two categories: static and dynamic elements. Layers 1 and 2 represent the static elements, while Layers 3, 4, and 5 define the dynamic elements. Consequently, we construct five axes within the search space: the static element axis, temporary element change axis, traffic participant axis, weather environment axis, and an additional axis that serves as an extension interface for other elements. Our knowledge-driven framework for the generation of safety-critical scenarios, \titlename, is built on this search space. Employing this framework, we extract parameters from each axis of the search space and synergize them, crafting discrete scenarios that are compatible with the Carla simulator.

\subsection{Constraints on Typical ODD Scenario Elements}

We identified that each layer of the ODD and the elements within these layers inherently possess constraints, as illustrated in Fig.~\ref{fig:Constraint relationship}. Neglecting these constraints can lead to the generation of scenarios that lack realism and might produce meaningless or ineffective testing scenarios. To enhance the authenticity of scenario generation and improve testing efficiency, we delved into the constraint relationships between ODD scenario elements. These relationships are categorized into inter-layer constraints, intra-layer constraints, and inter-element constraints. Note that there are more inherent constraints in typical ODD scenario elements. In this section, we present exemplary cases to explain and highlight their importance in maintaining the realism of the generated safety-critical scenarios.

\subsubsection{Inter-layer Constraints} 

\begin{itemize}
    \item \textit{Road Network Structure to Traffic Facilities (Layer1 $\rightarrow$ Layer2)}: The topology of Layer1's road network applies the placement constraints on Layer2's traffic facilities. For instance, traffic lights are only permitted to be initialized at intersections. Similarly, signposts or street lights can only be placed alongside roads, and their placement in the middle of the road or outside of it is restricted.
    
    \item \textit{Road Network Structure to Traffic Participants (Layer1 $\rightarrow$ Layer4)}: The design of Layer1's road structure imposes constraints on the behaviors of Layer4's traffic participants. For example, pedestrians are only allowed to move on crosswalks, and vehicles are only allowed to drive on roads. Furthermore, their speed and driving trajectories are constrained by the road network design.
    
    \item \textit{Traffic Participants to Temporary Changes (Layer4 $\rightarrow$ Layer3)}: The behavior of Layer4's traffic participants can influence temporary changes in Layer3. For instance, collisions and congestion among traffic participants can lead to traffic accidents and jams, resulting in temporary modifications in the simulated scenarios. Therefore, the behavior of traffic participants largely affects the occurrence of unexpected events.
    
    \item \textit{Weather Environment to Road Network Structure (Layer5 $\rightarrow$ Layer1)}: The weather conditions of Layer5 can influence the drivable areas of Layer1. For instance, during heavy fog or rain, the use of highways might be restricted.
    
    \item \textit{Weather Environment to Traffic Participants (Layer5 $\rightarrow$ Layer4)}: The weather conditions of Layer5 can influence the behavior of Layer4's traffic participants. For instance, during nighttime, heavy rain, or thick fog, the driving speed of traffic participants might be limited.
\end{itemize}

\subsubsection{Intra-layer Constraints}

For the intra-layer constraints, our primary focus is on the often-overlooked yet crucial constraints present in the environment layer, namely: cloud cover, rainfall, wind speed, fog density, and illumination~\cite{czarnecki2018operational,thorn2018framework}. In popular simulators like Carla~\cite{dosovitskiy2017carla}, these environmental components are treated as mutually independent, deviating from the realism observed in the natural world. Our objective is to rectify this oversight.

\begin{itemize}
    \item \textit{Rainfall constrains cloud cover}: High rainfall typically requires a higher amount of cloud cover.
    \item \textit{Wind speed constrains fog density}: When wind speed is high, fog density should be limited to lower values.
    \item \textit{Fog density and cloud cover constrain illumination}: As the thickness of fog and cloud cover increases, light absorption increases and light transmittance decreases. Therefore, when either fog density or cloud cover is high, illumination should be constrained to lower values.
\end{itemize}

\subsubsection{Inter-element Constraints}

Apart from the aforementioned constraints, we pay attention to the constraints between several significant elements.

\begin{itemize}
    \item \textit{Rainfall}: Rainfall has two internal element attributes: precipitation intensity and precipitation deposits. When precipitation intensity increases, the precipitation deposits increase as well. At the same time, these rainfall changes will influence wetness and friction, two internal elements of the road. As wetness increases, the friction decreases. In Carla, this relationship can be represented as
    \begin{equation}
        \text{friction} = \begin{cases}
            \frac{1-\text{wetness}}{200}, & \text{if wetness} < 40 \\
            0.6, & \text{if wetness} \geq 40
        \end{cases}
    \end{equation}
    \item \textit{Fog}: When fog density is high, fog falloff is high but the fog distance should be limited to lower values, as illustrated below. 
    \begin{equation}
        \begin{aligned}
            \text{fog}_\text{distance} & = 100 - \text{fog}_\text{density}, \\
            \text{fog}_\text{falloff} & = 0.05 \times \text{fog}_\text{density}
        \end{aligned}
    \end{equation}
    \item \textit{Illumination}: As the thickness of fog and cloud cover increases, light absorption increases, and light transmittance decreases. Therefore, when either fog density or cloud cover is high, illumination should be constrained to lower values.
\end{itemize}

Illumination has two internal attributes, altitude, and azimuth, and they are related in the way:
\begin{equation}
\begin{gathered}
    \text{azimuth} = \frac{125}{4} \arcsin \left(\frac{\text{altitude} + 20}{70}\right) \in \left[-\frac{125}{4}, \frac{125}{4}\right]
\end{gathered}
\end{equation}

\section{Ontology-based Autonomous Driving Scenario Modeling}\label{sec:Ontology}
Since knowledge is the recognition and understanding derived from different people's processing of information, difficulties in utilizing knowledge often arise from not expressing it accurately or effectively. By constraining and modeling ODD elements based on ontology, scenarios can be made readable by both humans and machines, thereby achieving structured scenario generation. This section describes how to accomplish the aforementioned content based on ontology.

\subsection{Tools for Ontology}
Humans are more adept at handling abstract knowledge, but this knowledge may be inaccessible to machines. Therefore, it is necessary to structuralize the knowledge. Ontology employs a symbol-based knowledge representation method called the resource description framework~(RDF), which can represent natural language as a triplet containing subject~(S), predicate~(P), and object~(O) for representation and storage. Protégé is a framework-based ontology editing and modeling tool, mainly used for class modeling, entity editing, object property, and data property definition, among other things~\cite{mealy1967another}. It uses the web ontology language~(OWL) to represent knowledge, and the represented knowledge can be easily understood and applied by machines. Therefore, we chose Protégé to build the ontology conceptual model of the driving scenario.

% figure 4
\begin{figure*}
    \centering
    \includegraphics[width=.8\linewidth]{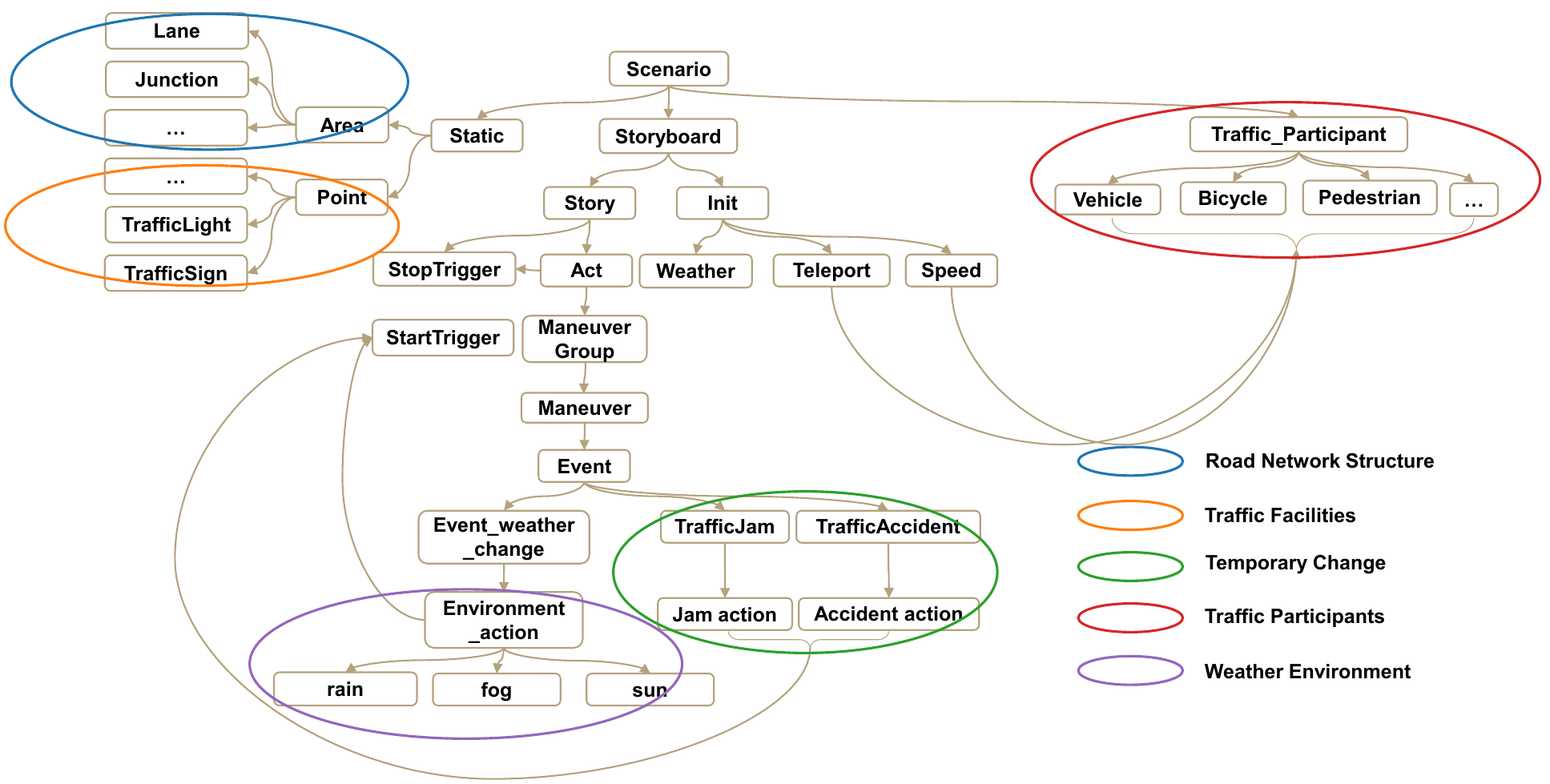}
    \caption{The Framework of Ontology Modeling.}
    \label{fig:Ontology Modeling Framework}
\end{figure*}

\subsection{Definitions}

\subsubsection{Class}

Classes describe different concepts within driving scenarios and map elements within the 5-layer structure of ODD. In automated driving scenarios, road network structures have set area and point classes. Traffic facilities classes represent a set of traffic facility entities encountered by automated vehicles, including ConstructionCard~(construction signs), RoadCone~(cones), TrafficLight~(traffic lights), Signage~(signs), etc. Temporary changes are composed of events, including TrafficAccident~(traffic accidents), TrafficJam~(congestion), etc. Traffic participants include Pedestrian~(pedestrians), Vehicle~(vehicles), Bicycle~(bicycles), and Misc~(others). Action classes represent the set of actions of traffic participants, including SpeedAction~(change speed), LaneChangeAction~(change lanes), and TeleportAction~(locate absolute world position), etc. Weather environment classes are used to describe the climate and environmental status in the scenario, specifically describing three classes: fog, rain, and sun.

\subsubsection{Object Property}

Object properties describe the relationships between classes, constraining the described relationships through Domain and Range, both of which are class types. Constraints across different scenario layers are also achieved through object properties. For example, the object property `hasSun' represents the interaction relationship within the weather environment and the relationship between weather and sunlight.

\subsubsection{Data Property}

Data properties describe the class and also constrain the described relationships through Domain and Range, where the Domain is a class type, and the Range is a basic data type. For example, traffic participants and traffic facilities have properties like has\_world\_x, has\_world\_y, has\_world\_z, has\_world\_pitch, has\_world\_yaw, has\_world\_roll to represent their position.

\subsubsection{Individual/Instance}

The individual or instance refers to the instantiation using the above-mentioned class and properties from practical driving scenarios.

\subsection{Ontology-based Modelling}

As shown in Fig.~\ref{fig:Ontology Modeling Framework}, the ontology modeling includes:

\begin{itemize}
    \item \textbf{Layer1 and Layer2}: By defining the AreaEntities, regional entities of the Static class, including intersection size, lane length, number of lanes, and other road topological structures, the modeling of Layer1 is achieved. By defining the PointEntities, point entities of the Static class, including traffic lights, traffic signs, etc., the modeling of Layer2 is achieved.
    \item \textbf{Layer3}: The implementation of temporary feature changes, including traffic accidents and congestion, needs to be based on the Event class. Corresponding changes are encapsulated into different events. These events consist of traffic participants~(Who) and storyboards~(What) and are triggered by Condition and Trigger~(When) to start the event.
    \item \textbf{Layer4}: Defines motor vehicles, non-motor vehicles, pedestrians, and other traffic participant classes, along with their actions, such as using Teleport and Speed to define their initial positions and speeds, and using SpeedAction, LaneChangeAction to define the modifications of traffic participant behaviors.
    \item \textbf{Layer5}: By setting EnvironmentAction, elements such as rainfall, fog volume, and illumination are connected to the Weather Environment class through the has\_weather data property. By utilizing EnvironmentChangeEvent, the modification of various Weather Environment attributes can be encapsulated into an event.
\end{itemize}

\subsection{XOSC File Generation Process Based on Ontology}

We further developed an ontology modeling tool based on Python, implementing the connections between various elements. Given the popularity of the OpenX standard~\cite{chen2022generating,bogdoll2022one}, we generate OpenDRIVE and OpenSCENARIO files that support most simulation executions. The specific process is shown in Algo.~\ref{alg:Ontology Modeling}.

\begin{algorithm}[t]
\caption{Ontology-based simulation scenarios generation}
\label{alg:Ontology Modeling}
\textbf{Input:} Hyperparameter vector X\\
\textbf{Output:} Files for static scenarios~(.xodr) and files for dynamic scenarios~(.xosc)
\begin{enumerate}
\item Using Protégé to construct an ontology-based template for the 5-layer ODD structure.
\item Static scenario modeling
\begin{enumerate}[label=\arabic*., leftmargin=*] % This changes the labels to numbers
\item Take the road network structure in Layer1 and the static parameter vector $X_{\text{static}}$ of the corresponding elements in Layer2 traffic facilities as input, and use the ontology modeling tool to generate static element owl files.
\item Use the OWL2OpenDRIVE tool to convert the static element owl file into a .xodr file as the output.
\end{enumerate}
\item Dynamic scenario modeling
\begin{enumerate}[label=\arabic*., leftmargin=*] % This changes the labels to numbers
\item Take the .xodr file as input, receive the dynamic parameter vector $X_{\text{dynamic}}$ corresponding to Layer3-Layer5, and use the ontology modeling tool to generate dynamic scenario owl files.
\item Use the OWL2OpenSCENARIO tool to convert the dynamic scenario owl file into a .xosc file as the output.
\end{enumerate}
\end{enumerate}
\end{algorithm}

\subsection{The Scenario Generation Algorithm Toolkit}

The generation algorithms used in our toolkit for ADV validation are categorized into:
\begin{itemize}
    \item {\bf Single-Objective Optimization Algorithms:} Such as PSO~\cite{kennedy1995particle}, GA~\cite{holland1992genetic}, and ES~\cite{beyer2002evolution}, suitable for problems with a single objective function.
    \item {\bf Multi-Objective Optimization Algorithms:} Including SPEA~\cite{zitzler2001spea2}, MOPSO~\cite{reyes2006multi}, and NSGA-II~\cite{deb2000fast}, used to balance different objectives.
    \item {\bf Deep Reinforcement Learning:} Algorithms like DQN~\cite{fan2020theoretical}, DDPG~\cite{barth2018distributed}, and PPO~\cite{schulman2017proximal}, designed for complex environments but require sufficient data.
\end{itemize}
The choice among these depends on the problem's characteristics, with traditional algorithms focusing on specific objectives and reinforcement learning adapting to complex interactions. The toolkit supports these algorithms for tailored selection.

% table 1
\begin{table}
\centering
\caption{Summary of Critical Metrics}
\label{tab:metrics}
\setlength{\tabcolsep}{10pt} % Adjust the column separation
\begin{tabular}{c|p{0.28\textwidth}}
\toprule
\textbf{Criticality Metric} & \textbf{Explanation} \\ 
\midrule
\(r_{\text{red}}\)  & Whether through a red light \\
\(n_{\text{collision}}\) & Number of collisions\\
\(TTC\) & Measures the proximity of vehicles\\
\(d_{\text{min}}\) & Minimum distance from traffic participants\\
\(v_d\) & \(d_{\text{min}}\) corresponds to the speed of the ego vehicle \\
\(l_{\text{offset}}\) & Lane offset \\
\(IoU\) & Evaluates target detection accuracy \\
\(e_{\text{position}}\) & Measures target detection position deviation \\
\(mAP\) & Considers accuracy and recall rates for each category \\
\(a_{\text{change}}\) & Acceleration change \\ 
\(n_{\text{brake}}\) & Number of sudden stops \\
\bottomrule
\end{tabular}
\end{table}

\subsection{Criticality Metrics}
\label{sec:metrics}

The critical metrics are designed to detect high-risk, boundary, and collision scenarios, serving as crucial objectives for the optimization toolkit. A summary of these metrics is provided in Table.~\ref{tab:metrics}. 

\begin{equation}
        r_{ij} = \frac{x_{ij}^{\prime} - \min(x_j^{\prime})}{\max(x_j^{\prime}) - \min(x_j^{\prime})}
        \label{eq:normalize}
    \end{equation}

    \begin{equation}
        E_j = -\frac{1}{\ln m}\sum_{i=1}^m p_{ij}\ln p_{ij}
        \label{eq:entroy}
    \end{equation}
    \begin{equation}
        p_{ij} = \frac{r_{ij}}{\sum_{j=1}^n r_{ij}}
        \label{eq:account for}
    \end{equation}

    \begin{equation}
        w_{ij} = \frac{(1 - E_j)}{\sum_{j=1}^n(1 - E_j)}
        \label{eq:cal weight}
    \end{equation}

To derive an objective function for scenario generation algorithms, we use a weighted calculation of various critical metrics. An entropy-based method~\cite{jiang2021hybrid} is employed to determine the weights, involving data positivization and standardization, entropy calculation, and weight determination. The process is described by Eqs.~\eqref{eq:normalize}, \eqref{eq:entroy}, and \eqref{eq:cal weight}.

\section{Evaluation}\label{sec:Experiment}
In this section, we present a comprehensive evaluation using the popular Carla simulator. Crossroads, as a representative driving scenario, are the focus of this analysis. The safety-critical scenarios were carefully considered by defining the hyperparameters and constraints. A specially designed cost function guides the generation and optimization~(both single-objective and multi-objective optimizations) of these test scenarios, benchmarked against a random sampling algorithm~(RS).

\subsection{Simulation Scenario Design}

We constructed a complex urban crossroad scenario, featuring the ego vehicle and a single background vehicle(BV) as the traffic participants. Essential parameters such as the starting and ending positions, as well as the speeds of both vehicles, are configured to emulate real-world driving conditions. The design of these scenarios consists of four distinct types of collision points, each representing a unique interaction within the crossroad.

% table 2
\begin{table}
\centering
% \caption{Scenario Specifications.}
\caption{Logical Scenario Design and Attribute Configuration}
\label{tab:scenarios}
\setlength{\tabcolsep}{3pt}
\begin{tabular}{c|c|c|c|c|c|c|c}
\toprule
\textbf{Scenario} & \textbf{S$_{\text{ego}}$} & \textbf{E$_{\text{ego}}$} & \textbf{V$_{\text{ego}}$} & \textbf{S$_{\text{BV}}$} & \textbf{E$_{\text{BV}}$} & \textbf{V$_{\text{BV}}$} & \textbf{C}\\ 
\midrule
S$_1$ & P$_5$ & P$_2$ & 6.9\,m/s & P$_3$ & P$_8$ & 1.8\,m/s & C$_1$ \\
S$_2$ & P$_5$ & P$_2$ & 6.9\,m/s & P$_1$ & P$_4$ & 1.8\,m/s & C$_2$ \\
S$_3$ & P$_1$ & P$_6$ & 6.9\,m/s & P$_7$ & P$_2$ & 5.5\,m/s & C$_3$ \\
S$_4$ & P$_5$ & P$_8$ & 6.9\,m/s & P$_3$ & P$_8$ & 1.8\,m/s & C$_4$ \\
\bottomrule
\end{tabular}
\end{table}

\begin{figure}
    \centering
    \includegraphics[width=.4\linewidth]{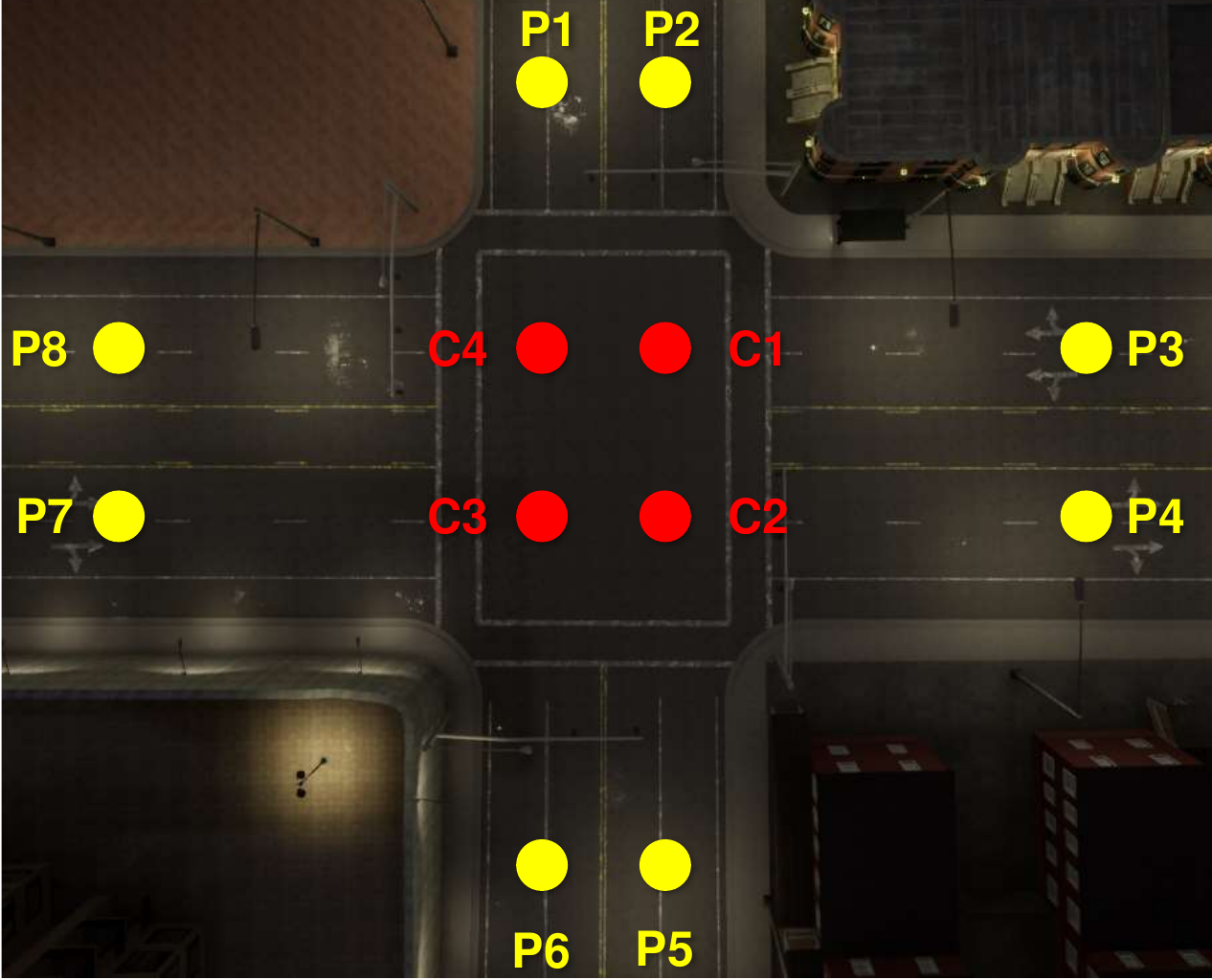}
    \caption{The crossroad scenario.@Carla}
    \label{fig:crossroads}
\end{figure}

\begin{figure}
    \centering
    \includegraphics[width=\linewidth]{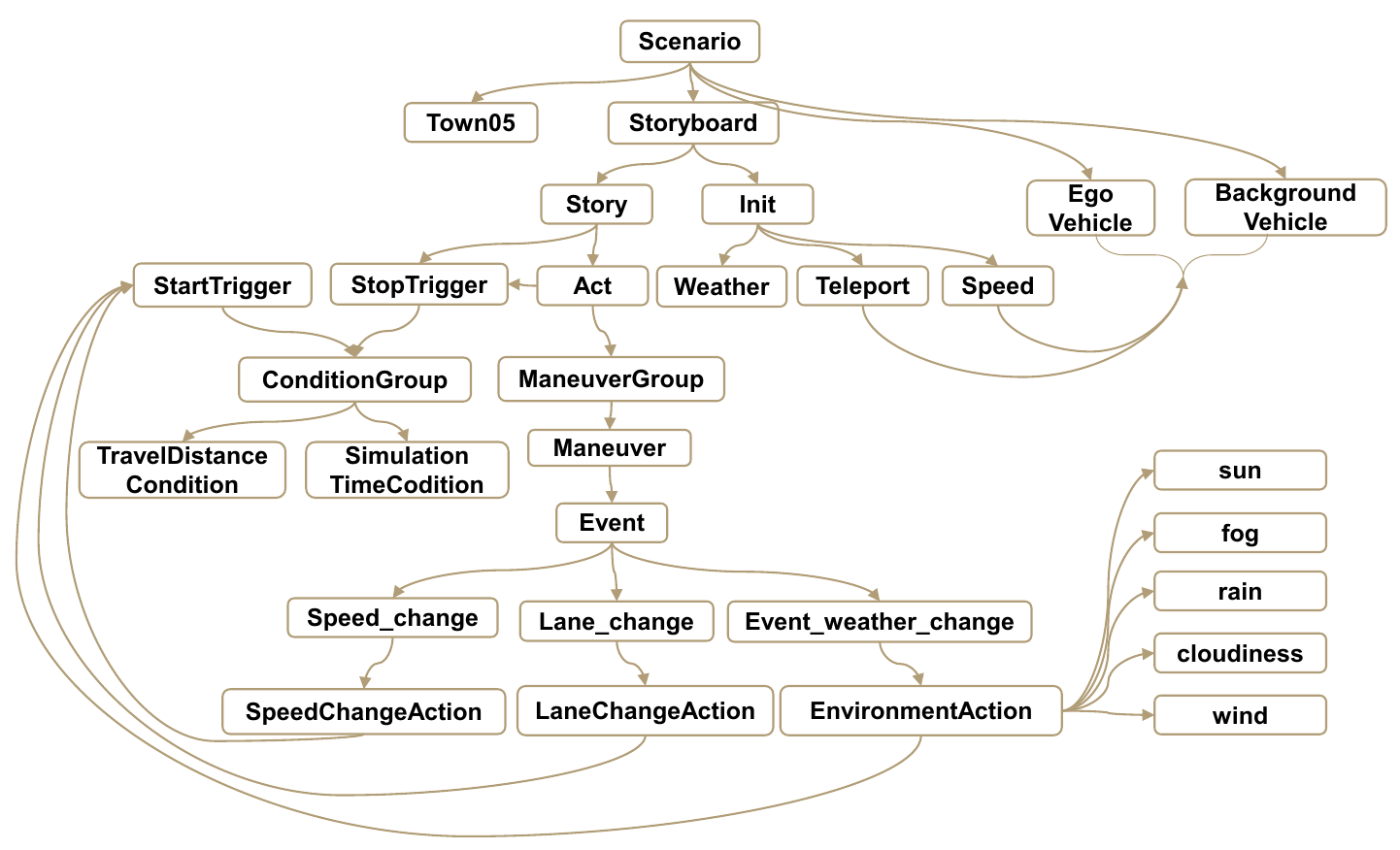}
    \caption{Ontology-Based Modeling of Experimental Scenarios.}
    \label{fig:Ontology modeling(Experiment)}
\end{figure}

As shown in Fig.~\ref{fig:crossroads}, four scenarios correspond to high-risk collision points $C_1$ to $C_4$, detailed in Table.~\ref{tab:scenarios}. We focus on the \(S_4\) scenario for detailed examination. In Carla, the ego vehicle adjusts its speed to \SI{2}{m/s}, enhancing realism and testing efficiency. The ontology-based method models the scenario (Fig.~\ref{fig:Ontology modeling(Experiment)}), using Carla's map Town05, with defined dynamic behaviors and triggers based on time or distance.

\subsection{The objective function and safety-critical scenarios.}

% Let \( v(t) \) and \( d(t) \) denote the VUT's velocity and distance to other vehicles at time \( t \), with critical moment \( t_0 \) given by \( d(t_0) = \min(d(t)) \). 
The objective function is \(\text{fitness} = a_1 \cdot d_{\min} + a_2 \cdot v_d\), with weights calculated using an entropy weighting method, yielding \( a_1 = 0.8297 \) and \( a_2 = 0.1703 \). Scenarios are divided based on criticality, with thresholds displayed in Table.~\ref{tab:Threshold Values}.

\begin{table}
\caption{Thresholds for Scenario Classifications}
\centering
\setlength\tabcolsep{18pt}  % adjusting table width (optional)
\begin{tabular}{c|c}
\toprule
    \textbf{Evaluation Metrics} & \textbf{Threshold for Critical Scenarios}   \\ 
\midrule
    $n_{collision}$ & $ > 0$ \\ 
    $d_{min}$ & $ < 2$ \\ 
    $v_d$  & $\geq 1$  \\
\bottomrule
\end{tabular}
\label{tab:Threshold Values}
\end{table}

% figure 5

% figure 6

% figure 7
\begin{figure}
    \centering
    \begin{subfigure}{.47\linewidth}
        \centering
        \includegraphics[width=\linewidth]{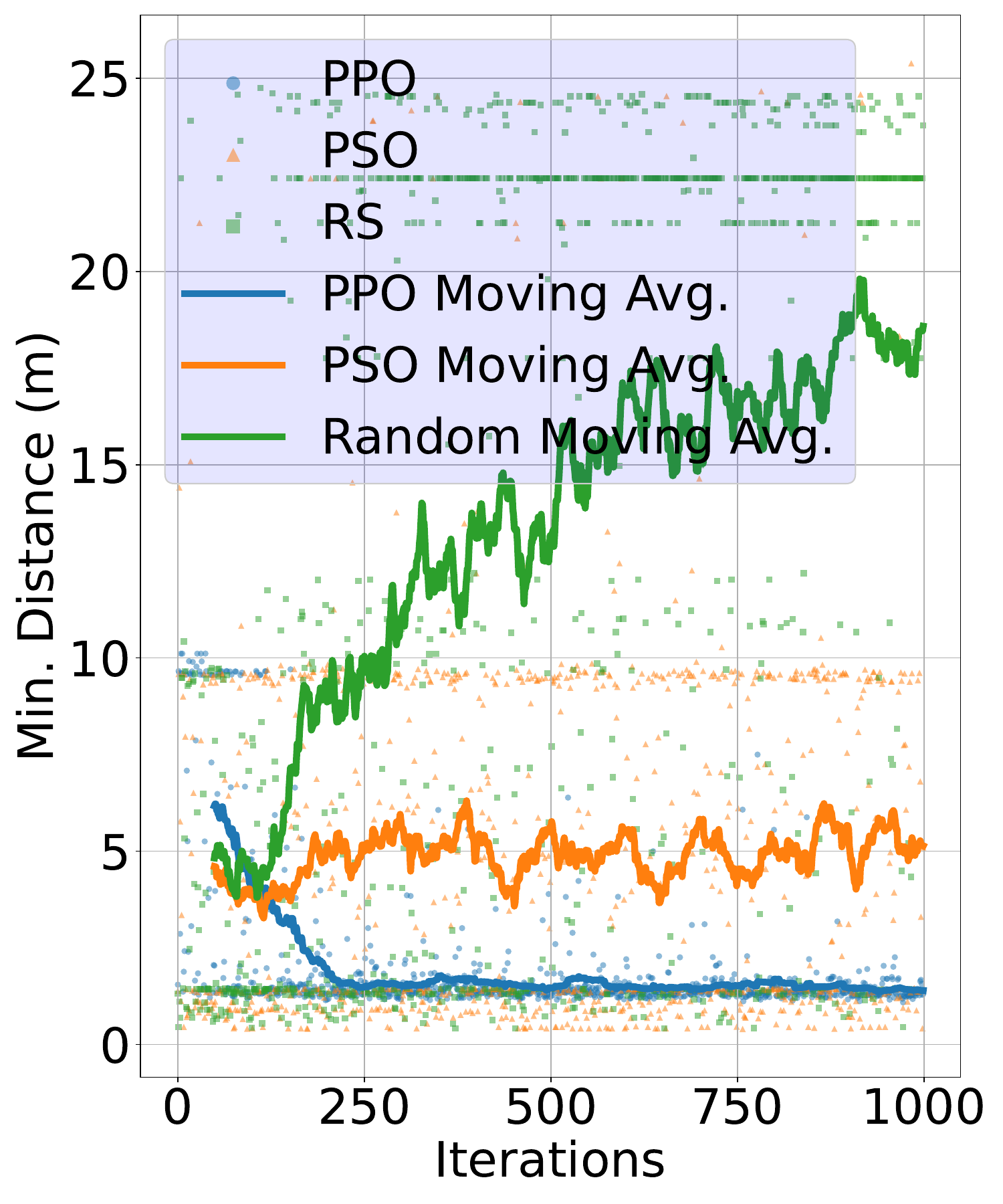}
        \caption{Min. Distance.}
    \end{subfigure}
    \begin{subfigure}{.47\linewidth}
        \centering
        \includegraphics[width=\linewidth]{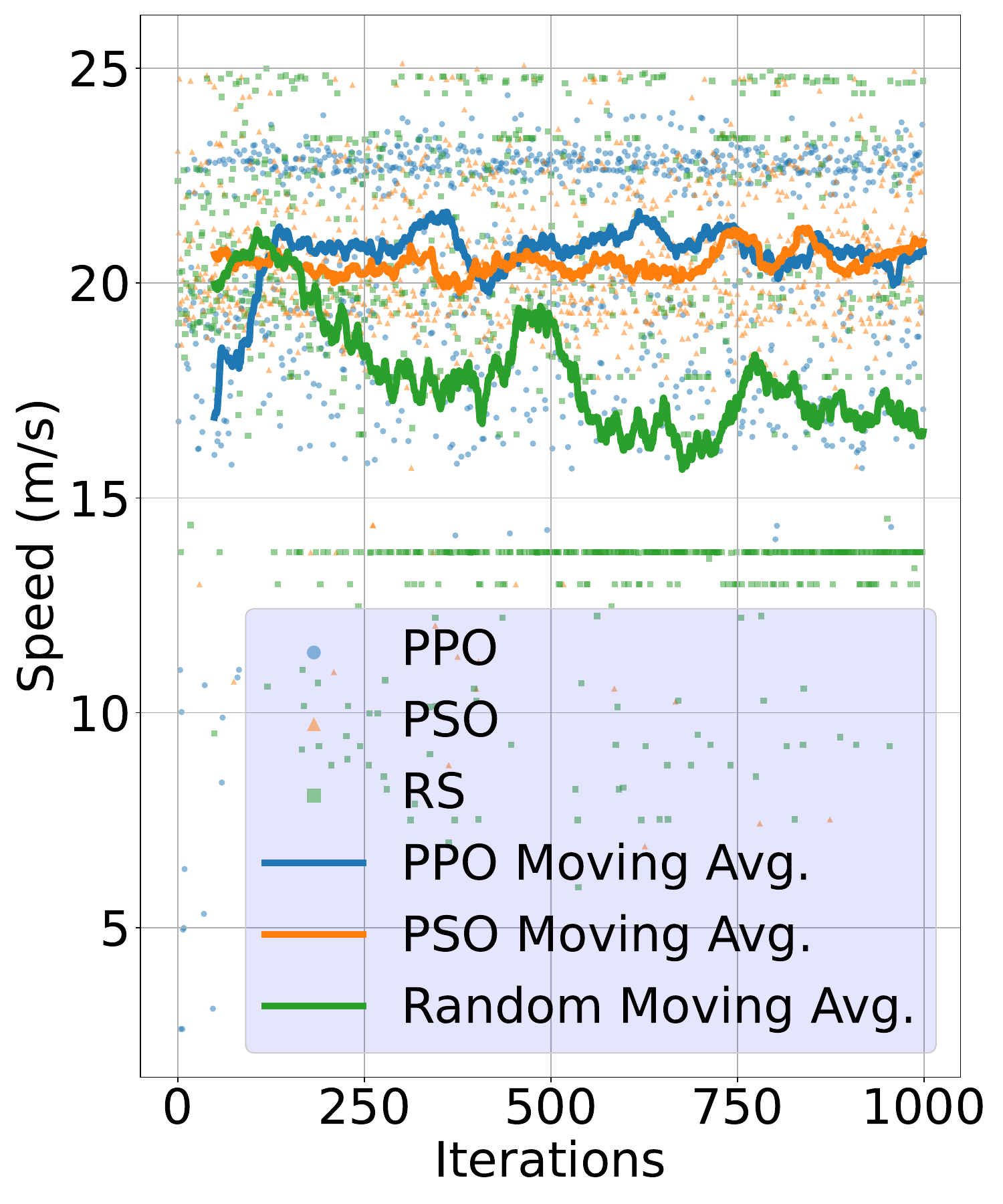}
        \caption{Speed.}
    \end{subfigure}
    \caption{The Distribution of Objective Attributes with Iterations in Single-Objective Optimization Algorithm.}
    \label{fig:single_opt_comp}
\end{figure}

% figure 8
\begin{figure*}
    \centering
    \begin{subfigure}{.32\linewidth}
        \centering
        \includegraphics[width=\linewidth]{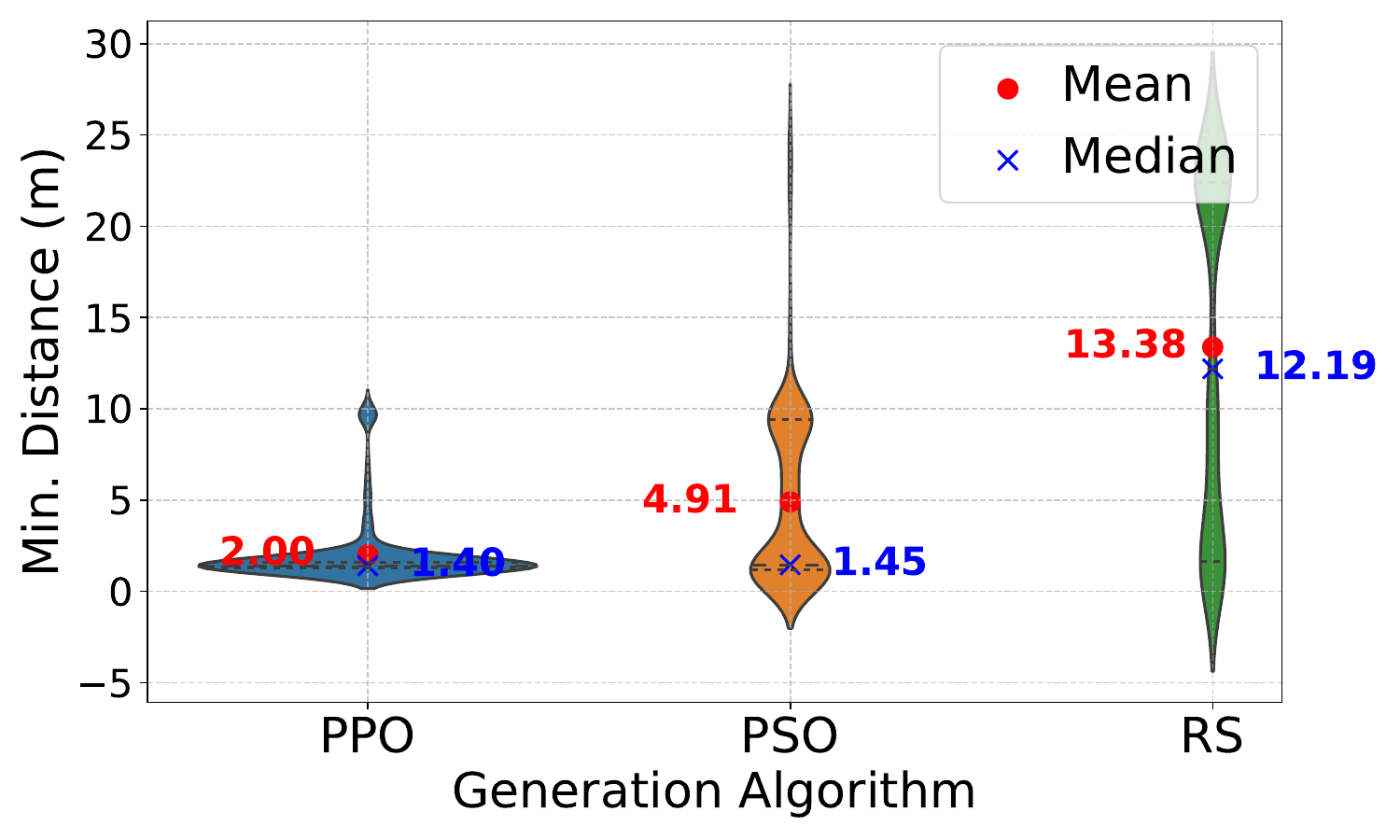}
        \caption{Min. Distance.}
    \end{subfigure}
    \begin{subfigure}{.32\linewidth}
        \centering
        \includegraphics[width=\linewidth]{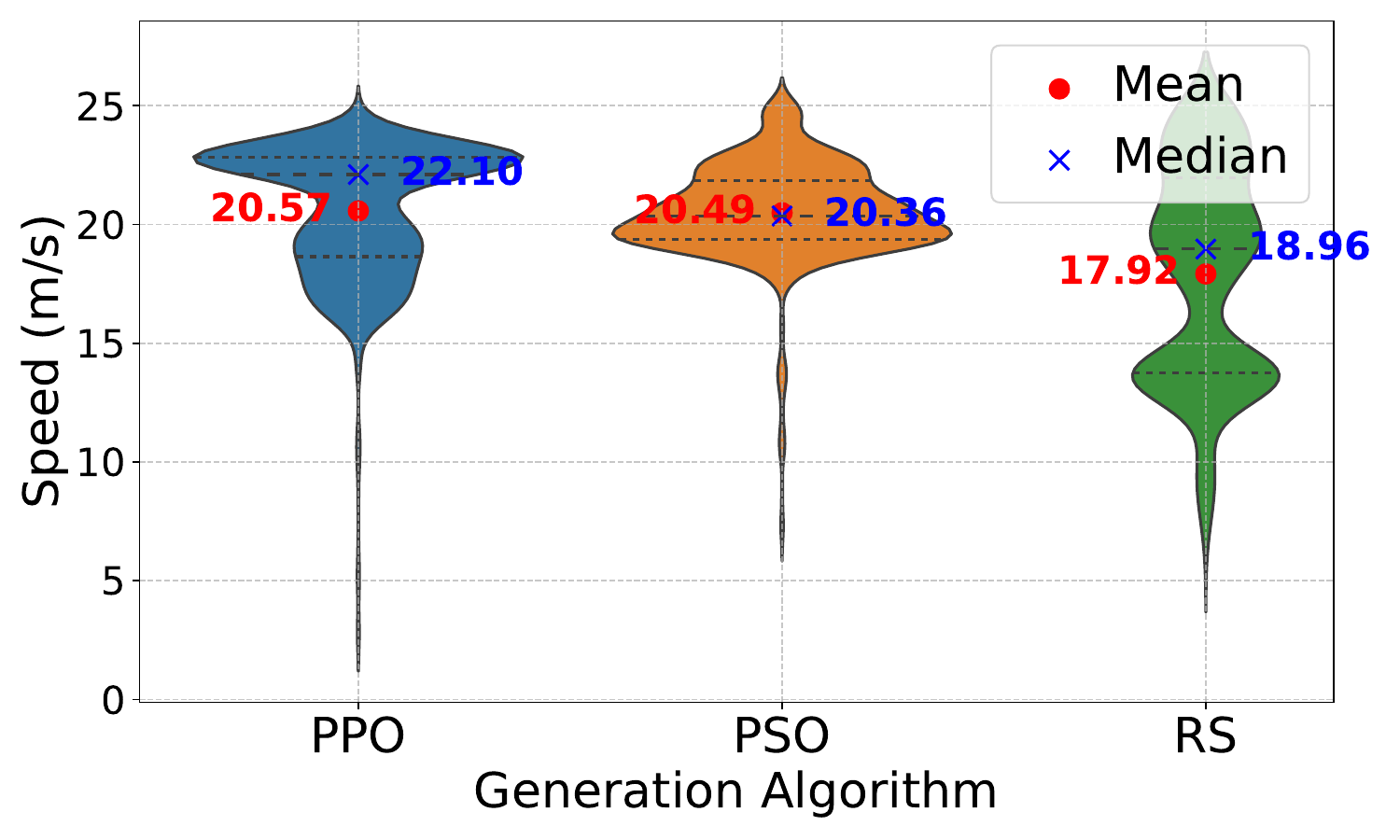}
        \caption{Speed.}
    \end{subfigure}
    \begin{subfigure}{.32\linewidth}
        \centering
        \includegraphics[width=\linewidth]{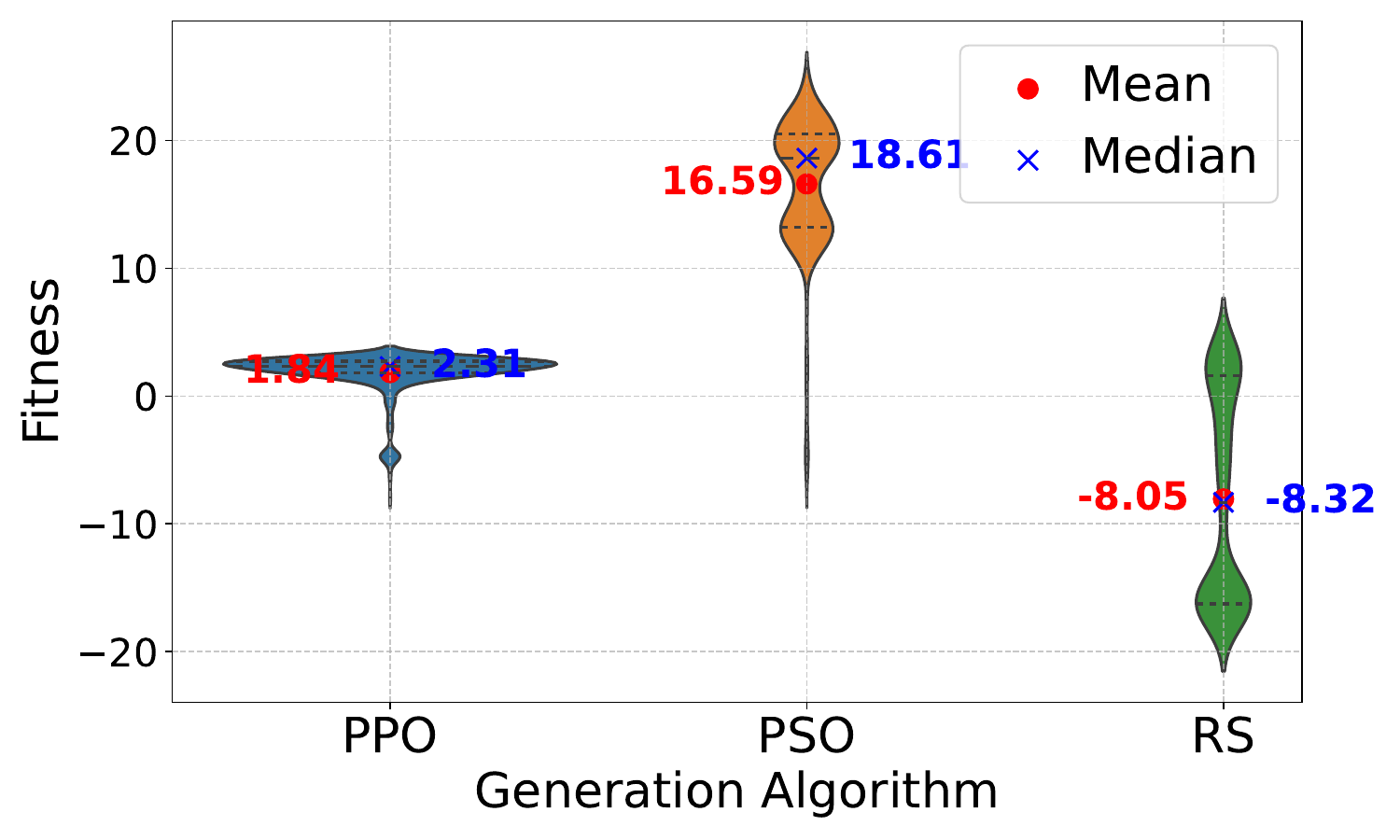}
        \caption{Fitness.}
    \end{subfigure}
    \caption{Violin plot of single-objective optimization algorithm evaluation index and objective function value.}
    \label{fig:single_distribution}
\end{figure*}

% figure 9
\begin{figure}[t]
    \centering
    \begin{subfigure}{.42\linewidth}
        \centering
        \includegraphics[width=\linewidth]{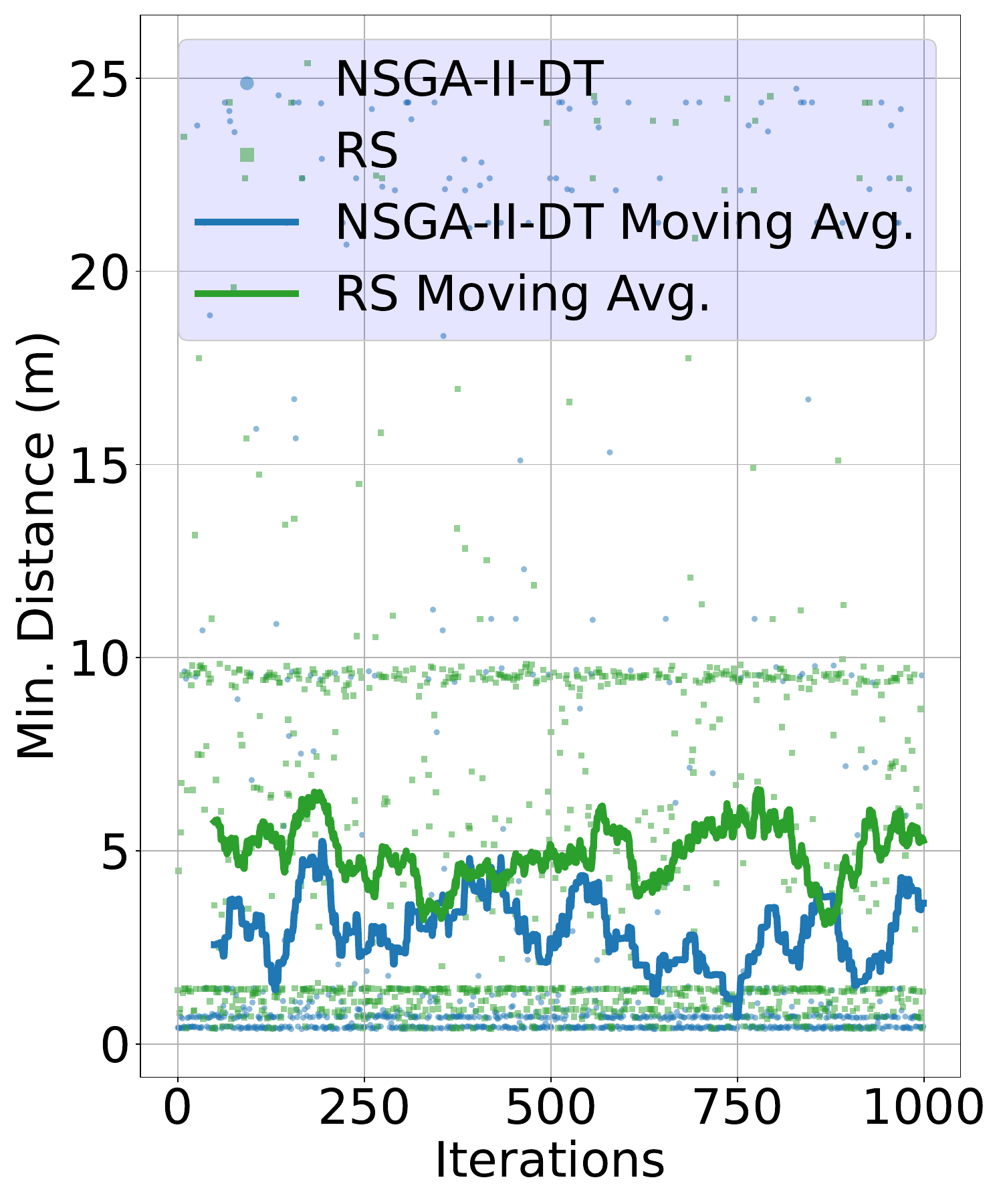}
        \caption{Min. Distance.}
    \end{subfigure}
    \begin{subfigure}{.42\linewidth}
        \centering
        \includegraphics[width=\linewidth]{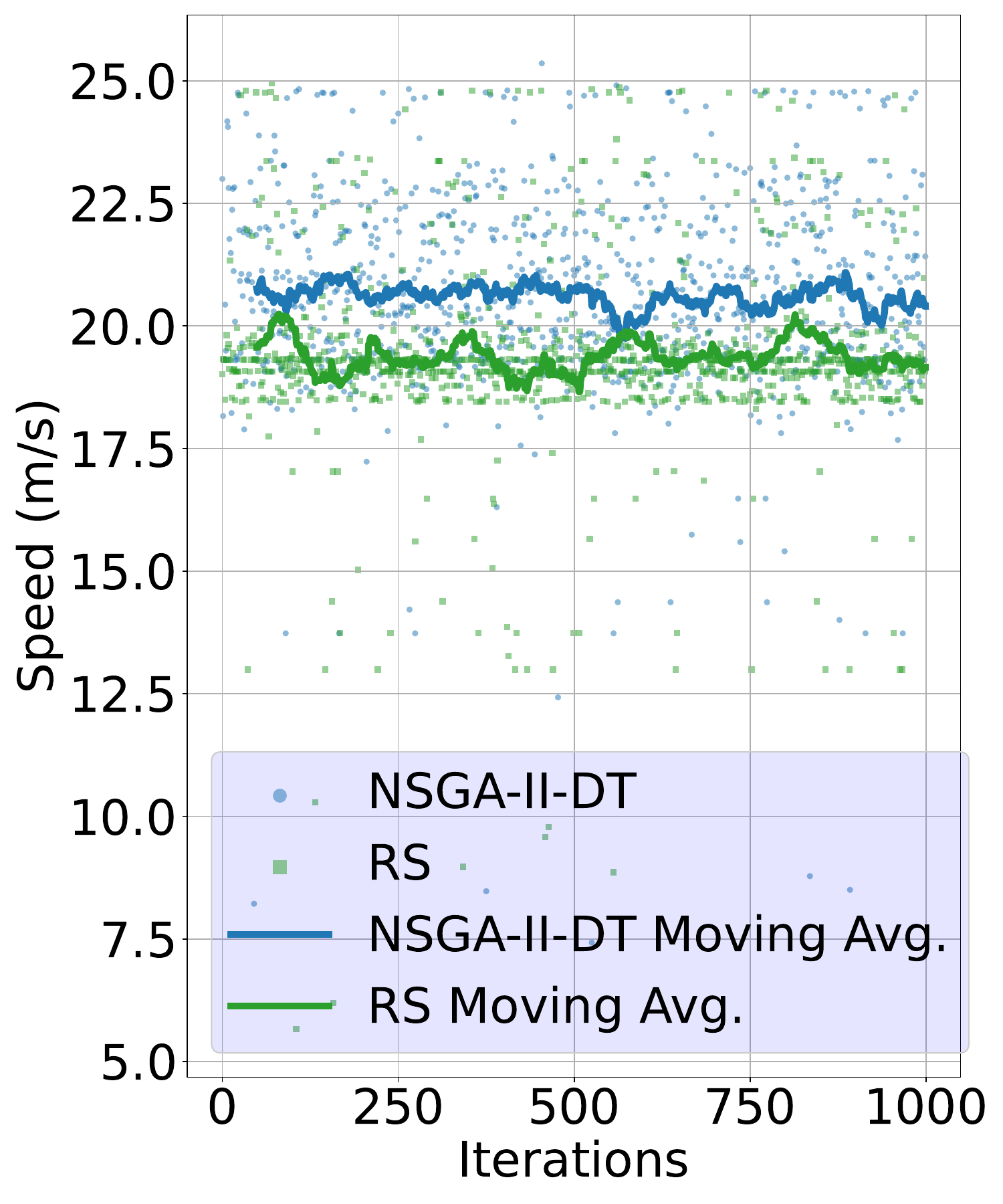}
        \caption{Speed.}
    \end{subfigure}
    \caption{The Distribution of Objective Attributes with Iterations in Muti-Objective Optimization Algorithm.}
    \label{fig:multiple_opt_comp}
\end{figure}

% figure 8
\begin{figure}
    \centering
    \begin{subfigure}{.45\linewidth}
        \centering
        \includegraphics[width=\linewidth]{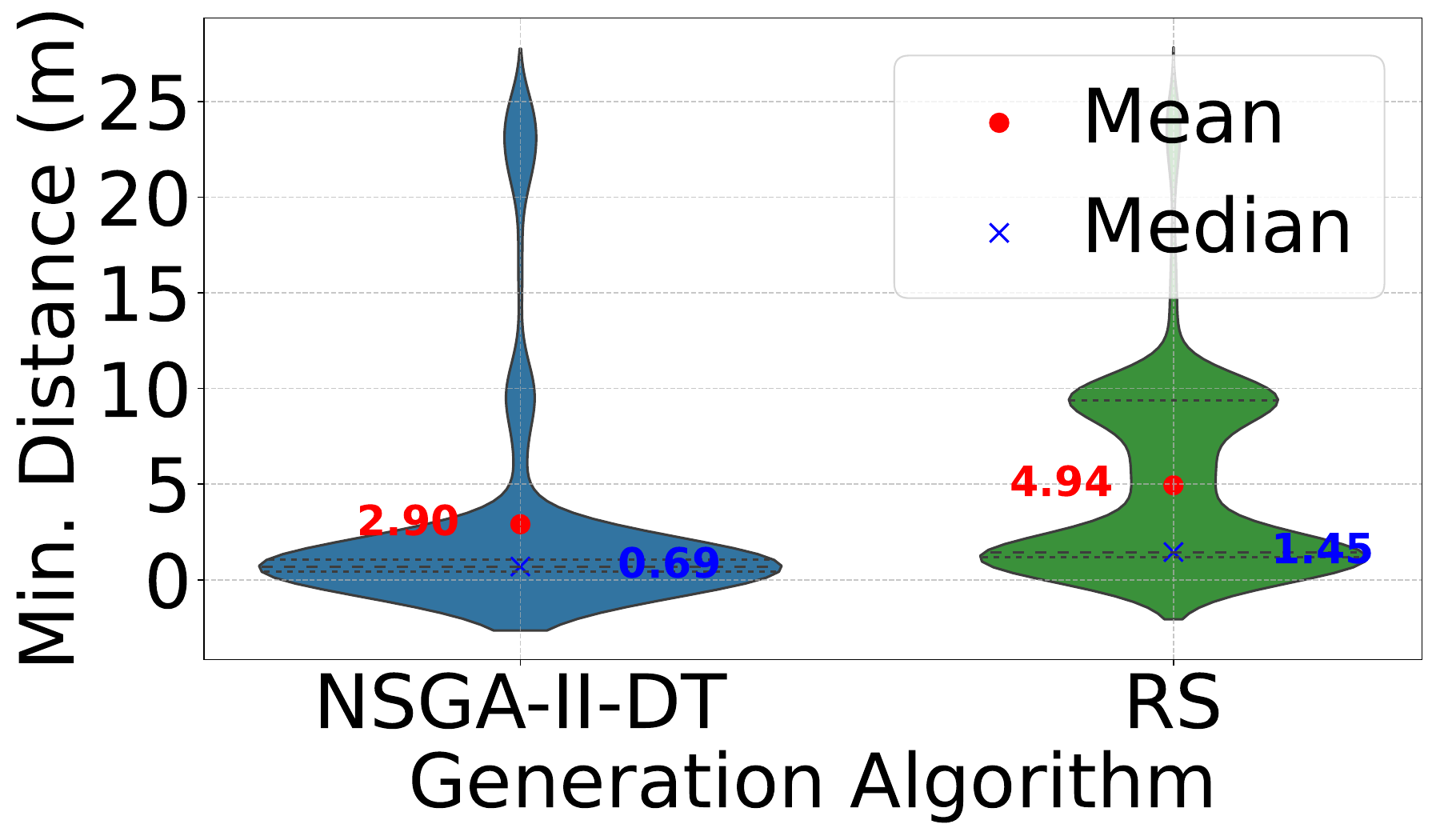}
        \caption{Min. Distance.}
    \end{subfigure}
    % \hfill
    \begin{subfigure}{.45\linewidth}
        \centering
        \includegraphics[width=\linewidth]{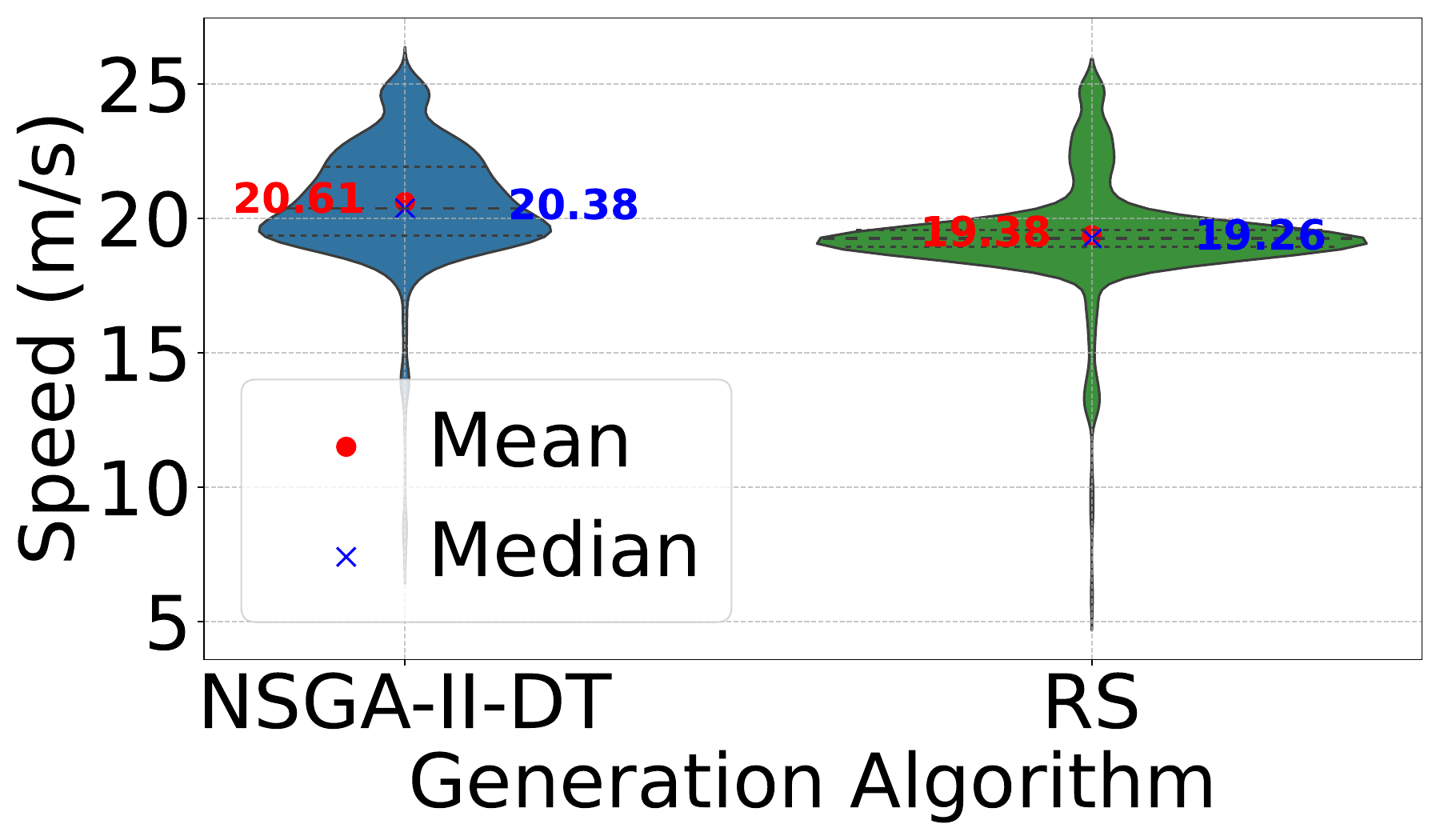}
        \caption{Speed.}
    \end{subfigure}
    \caption{Violin plot of the objective function of a multi-objective generative algorithm.}
    \label{fig:multi_distribution}
\end{figure}

\begin{figure}[t]
    \centering
    \includegraphics[width=\linewidth]{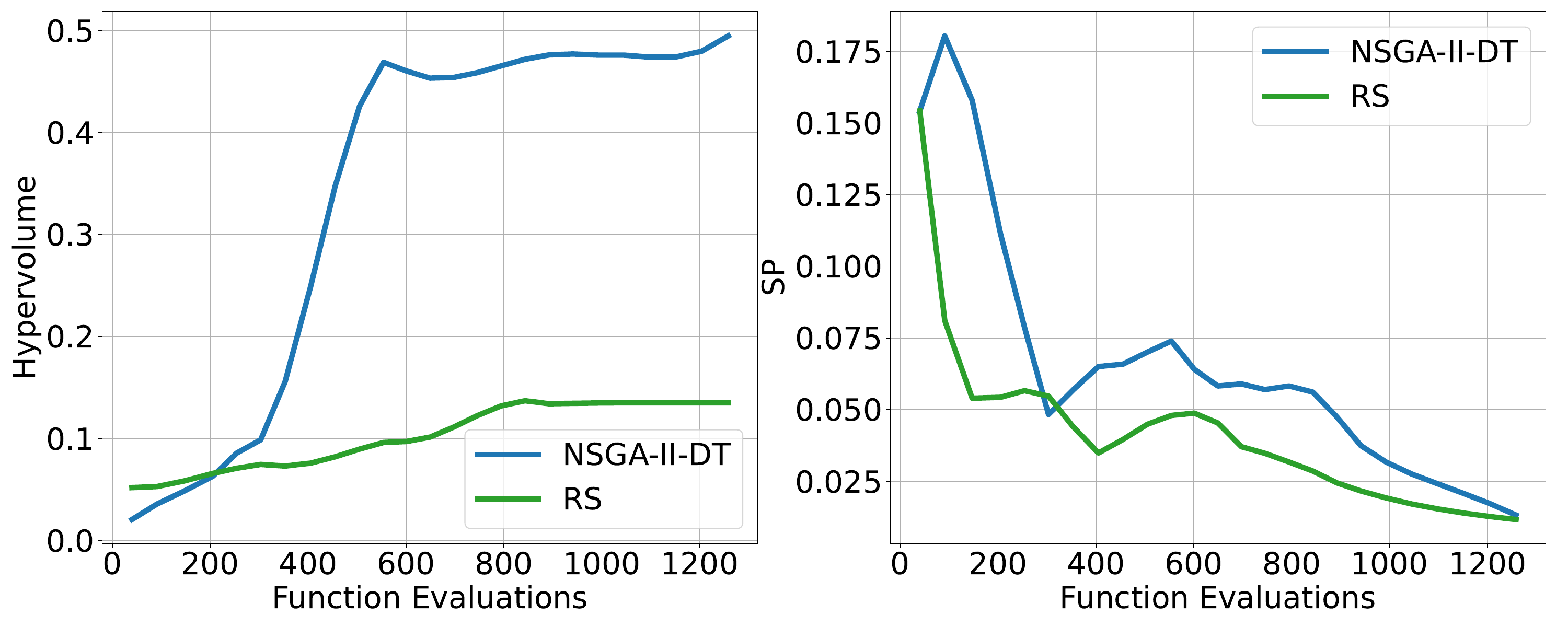}
    \caption{HV and Spread curves analysis.}
    \label{fig:curves}
\end{figure}

% table 4
\begin{table}
\caption{The Hyperparameter of Single-Objective Optimization Algorithms}
\centering
\setlength\tabcolsep{15pt}
\begin{tabular}{c|c|c|c}
\toprule
 & {\bf RS} & {\bf PPO} & {\bf PSO} \\
\midrule
Iterations & $25$ & $25$ & $25$ \\
Populations & $40$ & $40$ & $40$ \\
K\_epochs & $/$ & $32$ & $/$ \\
Gamma & $/$ & $0.99$ & $/$ \\
lr\_actor & $/$ & $0.0003$ & $/$ \\
lr\_critic & $/$ & $0.001$ & $/$ \\
$C_1$ & $/$ & $/$ & $1.5$ \\
$C_2$ & $/$ & $/$ & $1.5$ \\
W & $/$ & $/$ & 0.8 \\
\bottomrule
\end{tabular}
\label{tab:Hyperparameter(Single)}
% ... rest of the table remains the same
\end{table}

% table 5
\begin{table}
\caption{The Hyperparameter of Muti-Objective Optimization Algorithms}
\centering
\setlength\tabcolsep{20pt}
\begin{tabular}{c|c|c}
\toprule
 & {\bf RS} & {\bf NSGA-\uppercase\expandafter{\romannumeral2}-DT} \\
\midrule
    Iterations & $25$ & $25$ \\
    Population & $40$ & $40$ \\
    Crossover probability & $/$ & $0.7$ \\
    Mutation probability & $/$ & $0.05$ \\
\bottomrule
\end{tabular}
\label{tab:Hyperparameter(Muti)}
\end{table}

\subsection{Generative Algorithm Analysis}

The validation process encompasses the utilization of reinforcement learning and conventional single-objective and multi-objective optimization algorithms. This evaluation involves various scenarios, including varying numbers of objectives and the presence or absence of mathematical model training. Specific hyperparameters for each algorithm can be found in Table~\ref{tab:Hyperparameter(Single)} and Table~\ref{tab:Hyperparameter(Muti)}.

\subsubsection{Single-objective Scenario Generations}

From Table.~\ref{tab_3}, we observe critical scenario identification ratios($R_{critic}$) RS:PPO:PSO \(= 0.530:0.852:0.746\), and critical scenario generation time($T_{critic}$) RS:PPO:PSO \(= 5.098:1.000:1.954\), indicating superior efficiency of PPO and PSO over RS. Fig.~\ref{fig:single_opt_comp} and Fig.~\ref{fig:single_distribution} illustrate optimization outcomes and data distributions, showing PPO and PSO outperforming RS. Time analysis reveals no significant difference among the algorithms, with times RS:PPO:PSO = 6:46':6:06':6:28'. PPO is found to be more efficient but costlier in time, and both PPO and PSO improve critical scenario generation and optimization compared to RS.

\subsubsection{Multi-objective Scenario Generations}

In the evaluation of safety-critical scenarios, NSGA-II-DT outperforms RS in both efficiency and quality as shown in Table.~\ref{tab_3}. Figures~\ref{fig:multiple_opt_comp}, ~\ref{fig:multi_distribution}, and the HV and spread curves in Fig.~\ref{fig:curves} further confirm NSGA-II-DT's superior performance, due to its intrinsic need to balance two objective attributes. Though NSGA-II-DT required less runtime (7:04') compared to RS (7:47'), its computational time was longer than single-objective algorithms due to the complexity of multi-objective optimization. In summary, NSGA-II-DT demonstrates enhanced generation of critical scenarios across various metrics.

\begin{table}
    \centering
    \caption{{Experimental Results}}
    \setlength{\tabcolsep}{4pt}
    \renewcommand{\arraystretch}{1.2} % Default value: 1
    \begin{tabular}{cccccc} % Using the m column type for the first and last columns
        \toprule
        \multirow{2}{*}[-.5ex]{{\bf Method}} & \multicolumn{3}{c}{{\bf \makecell{Single-Objective \\Optimization}}} & \multicolumn{2}{c}{{\bf \makecell{Muti-Objective \\ Optimization}}} \\ % Multirow with vertical adjustment
        \cmidrule(lr){2-4} \cmidrule(lr){5-6} % Partial horizontal line
        & {\bf RS} & {\bf PPO} & {\bf PSO} & {\bf RS} & {\bf NSGA-\uppercase\expandafter{\romannumeral2}-DT} \\
        \midrule
        $R_{critic}$ & $0.530$ & $0.852$ & $0.746$  & $0.541$ & $0.649$ \\
        $T_{critic}$ & $5.089$ & $1.000$ & $1.954$ & $1.567$ & $1.000$\\
        $Avg(d_{min})$ & $13.38$ & $2.00$ & $4.91$ & $4.94$ & $2.90$\\
        $Median(d_{min})$ & $12.19$ & $1.40$ & $1.45$ & $1.45$ & $0.69$\\
        $Avg(v_d)$ & $17.92$ & $20.57$ & $20.49$ & $19.38$ & $20.61$ \\
        $Median(v_d)$ & $18.96$ & $22.10$ & $20.36$ & $19.26$ & $20.38$\\
        $Avg(fitness)$ & $-8.05$ & $1.84$ & $16.59$ & / & /\\
        $Median(fitness)$ & $-8.32$ & $2.31$ & $18.61$ & / & /\\
        $Runtime$ & $6:46'$ & $6:06'$ & $6:28'$ & $7:47'$ & $7:04'$ \\
        \bottomrule
    \end{tabular}
    \label{tab_3}
\end{table}

\section*{Acknowledgement}
\textcolor{blue}{This work was supported by the National Natural Science Foundation of China under Grants No.62232008.}

\section{Conclusion and Outlook}\label{sec:Conclusion}

This research presents a framework for generating safety-critical scenarios for testing ADS, with contributions including human- and machine-readable scenario files, enhanced efficiency in scenario generation, and comparative analysis of single- and multi-objective optimization algorithms. \textcolor{blue}{For future research, while our preliminary comparison with random sampling highlights the efficiency of our solution, a more comprehensive comparison with current work in the field could provide a detailed analysis of our approach's unique features. Additionally, subsequent development could focus on enhancing the completeness of the proposed modules.}

\bibliographystyle{IEEEtran}
\bibliography{sample-base}

\end{document}